# Value-Based Argumentation Frameworks

*Trevor Bench-Capon*
*Department of Computer Science*
*The University of Liverpool,*
*Liverpool,*
*UK*

Abstract

*In this paper I introduce the notion of Value-Based Argumentation Frameworks (VAF). I start from a standard notion of argumentation frameworks which has been widely used to analyse logics for defeasible argument, and extend it so as to make defeat dependent on the relative importance of the values the arguments advance or protect. The idea is to re-introduce an element which has been abstracted away in standard argumentation frameworks, and which can be used to ground a rational choice between alternatives which are equally tenable from the more abstract point of view.*

*The key result of the paper is to show that in a VAF it may be possible to force rational acceptance of particular arguments within the VAF,* irrespective of how the values are ranked. *Such arguments can be seen as objectively acceptable. In addition to the definition of VAFs, and exploration of their properties, particularly for the special, but rather common, case where there are precisely two values involved, the paper gives a detailed analysis of a well known moral dispute in terms of a VAF, and presents some heuristics for extending a given VAF to change the status of particular arguments.*

## 1. Introduction

In many of the more interesting situations where humans are called upon to reason it seems that finding an argument is not enough. For there are often many arguments pertinent to the situation which conflict with and support one another in a variety of ways. Instances of this abound in the context of moral and legal argumentation. In recognition of this, in recent years there has being a growing interest in *logics for defeasible argumentation* (e.g. Dung 1995, Bonderenko et al 1997). Such logics allow us to analyse situations in which we have arguments that may be overturned by other arguments, to see what, if anything, may safely be concluded. In particular the notion of an *Argumentation Framework* (*AF*), introduced by Dung, allows us to explore a system of conflicting arguments, in which arguments are abstracted to entities whose role is solely determined by their relation to other arguments. In moral and legal debate, however, an argument may attack another argument without defeating it. Suppose two people are debating whether income tax should be raised. One argues that it should because to do so would promote social equality, while the other argues that it should not, since enterprise should be rewarded. Certainly these arguments conflict, but neither party is likely to withdraw his argument in the face of the other. The point is that one believes that it is more important to promote equality than to reward enterprise, and the other ranks these purposes differently. Note also that both may accept *both* arguments: the first party can accept that it is right to reward enterprise, *but not at the expense of equality*. The point is that in many contexts the soundness of an argument is not the only consideration: arguments also have a force which derives from the value they advance or protect. Thus an argument may be defended not only by defeating its attacker, but also by ranking its value more highly than that of its attacker. That moral and legal disagreement should be seen in terms of different preferences for the values which the conflicting arguments defend or promote is the major insight of work in jurisprudence such as that of Perelman (e.g. 1980). If we want to model this kind of debate, the notion of an argument found in *AFs* is too abstract: we need also be able to relate arguments to their supporting values, and to allow these values to be ranked to reflect the preferences of the audience to which the arguments are addressed.

Accordingly, in this paper I extend Dung's *AF* into a *Value-Based Argumentation Framework* (*VAF*), by allowing this information to be represented. *VAFs* are defined in Section 2. When we have this notion, we can see that *VAFs* have some different properties from standard *AFs*, which can prove useful in the resolution of disputes. Important among these properties is that some arguments in a *VAF* will be in the preferred extension *independent of how the values are ranked*. These properties are discussed in Section 3. In Section 4 the use of a *VAF* is illustrated through an analysis of a well known problem from moral philosophy. Typically on entering a dispute a person will have some argument or arguments which they wish to be accepted. If the *VAF* does not produce this result, they will wish to extend it so that it does. Section 5 presents some strategic heuristics which will identify the kind of arguments with which they can usefully extend the *VAF*. Finally in Section 6 offers some concluding remarks.

## 2. Value-Based Argumentation Frameworks

Dung's definition of an Argumentation Framework (*AF*), given in Dung (1995), is:
**Definition 1**: An *argumentation framework* is a pair
$$AF = <AR, attacks>$$
Where *AR* is a set of arguments and *attacks* is a binary relation on AR, i.e. $attacks \subseteq AR \times AR$. For two arguments *A* and *B*, the meaning of *attacks(A,B)* is that *A* represents an attack on *B*. We also say that a set of arguments *S* attacks an argument *B* if *B* is attacked by an argument in *S*.

In order to represent the values to which arguments relate and the ranking of the values we must extend this definition to a Value-Based Argument Framework (*VAF*).
**Definition 2:** A *value-based argumentation framework* (*VAF*) is a 5-tuple:
$$VAF = <AR, attacks, V, val, valpref>$$
Where *AR*, and *attacks* are as for a standard argumentation framework, *V* is a non-empty set of values, *val* is a function which maps from elements of *AR* to elements of *V*, and *valpref* is a preference relation (transitive, irreflexive and asymmetric) on $V \times V$. We say that an argument *A* relates to value *v* if accepting *A* promotes or defends *v*: the value in question is given by *val(A)*. For every $A \in AF$, $val(A) \in V$.

Our purpose in extending the *AF* was to allow us to distinguish between one argument attacking another, and that attack succeeding, so that the attacked argument is defeated. We therefore define the notion of *defeat:*
**Definition 3:** *An argument $A \in AF$ defeats an argument $B \in AF$ if and only if both attacks(A,B) and not valpref(val(B),val(A)).*
Note that an attack succeeds if both arguments relate to the same value, or if no preference between the values has been defined. If *V* contains a single value, the *VAF* becomes a standard *AF*. If each argument maps to a different value, we have a Preference Based Argument Framework (Amgoud and Cayroll 1998).

Dung (1997) introduces the important notions *acceptability, conflict free set, admissible set, preferred extension* and *stable extension* for *AFs*. We also need to define these notions for a *VAF*.
**Definition 4**: An argument $A \in AR$ is *acceptable* with respect to set of arguments *S*, *(acceptable(A,S))* if:
$$(\forall x)((x \in AR\ \&\ defeats(x,A)) \rightarrow (\exists y)((y \in S)\ \&\ defeats(y,x))).$$
**Definition 5**: A set *S* of arguments is *conflict-free* if
$$(\forall x)(\forall y)((x \in S\ \&\ y \in S) \rightarrow (\neg attacks(x,y) \lor valpref(val(y),val(x)))).$$
**Definition 6**: A conflict-free set of arguments *S* is *admissible* if
$$(\forall x)(x \in S \rightarrow acceptable(x,S)).$$
**Definition 7:** A set of arguments *S* in an argumentation framework *AF* is a *preferred extension* if it is a maximal (with respect to set inclusion) admissible set of *AR*.
**Definition 8:** A conflict-free set of arguments *S* is a *stable extension* if and only if *S* attacks each argument in *AR* which does not belong to *S*.

The significance of these notions is with reference to which arguments in the *VAF* it is possible consistently to accept. If an argument is acceptable with respect to *AR*, it cannot be defeated and thus must be accepted. Similarly if an argument is acceptable with respect to a subset *S* of *AR*, it is possible to accept it, provided *S* is accepted. *S* cannot, of course, be consistently accepted unless it is conflict free. Thus an admissible set is one that can be consistently accepted given a *VAF*. If, however, an admissible set can be further extended, it is possible to accept further arguments. Thus a preferred extension represents a set of acceptable arguments to which no more arguments can be added. For standard *AFs* it is known that an *AF* always has a preferred extension (although it may be the empty set), and that preferred extensions are not in general unique. This last point means that it is possible for two parties to a dispute to adopt different positions which cannot be resolved by the *AF*. A stable extension cannot be empty: hence not every preferred extension is stable, and an *AF* may have no stable extension. Stable extensions are also not, in general, unique. In section 3 we will consider if these properties also hold true of a *VAF*.

We can now relate these notions to the ideas of credulous and sceptical acceptance in non-monotonic logic. An argument is *credulously acceptable* it appears in *at least one* preferred extension, and *sceptically acceptable* if it appears in *every* preferred extension.

Given a dispute, the ideal situation is that everyone concerned comes to accept precisely the same arguments. This requires that the preferred extension be unique. We say such a dispute is *resoluble*. Failing this ideal situation, we want to identify the consensus between the parties. This is represented by the sceptically acceptable arguments. If, however, an argument is not in this consensus, one of the parties may wish to show that it is at least defensible. For this to be so it must be credulously acceptable. We therefore say that an argument is *defensible* if it appears in at least one preferred extension, and *indefensible* otherwise. For a standard *AF* determining such things is not, in general easy. For example determining whether a dispute is resoluble is Co-NP complete, credulous acceptance is NP-complete, and sceptical acceptance even harder. See Dunne and Bench-Capon (2001) for a discussion of these and other similar results. The next Section will explore how these properties are affected by moving to a *VAF*.

## 3. Properties of *VAFs*

A useful way to picture an *AF* is as a directed graph in which the arguments are vertices and the attacks are represented as edges, directed from attacker to attacked. For a *VAF* we may picture the nodes being coloured differently to represent different values. Most of the complexity problems with *AFs* arise from the possibility of cycles in such a graph. Consider first a cycle with an odd number of vertices, say a three cycle. Such a cycle will have an empty preferred extension: If we suppose a node to be in the preferred extension its successor will be defeated. But this means that the third argument will not be defeated, and this will then defeat the argument that we originally supposed to be in the preferred extension. If, on the other hand, we consider a cycle with an even number of vertices, we will get two disjoint preferred extensions. Let us now refer to cycles in a *VAF* as *monochromatic* if they contain arguments relating to a single value, *dichromatic* if they contain arguments relating to exactly two values, and *polychromatic* if they have two or more values.

In a *VAF*, monochromatic cycles of odd length (*odd-cycles*) will have the empty set as their preferred extension and monochromatic cycles of even length (*even cycles*) will have two preferred extensions, as in a standard *AF*. We should, however, feel somewhat uncomfortable with the existence of monochromatic cycles in our graph. An odd cycle will have the nature of a paradox, in that we cannot consistently accept anything, and an even cycle the nature of a dilemma, in that we must choose between two alternatives on no rational grounds. Note that this criticism does not apply to cycles in a standard *AF*: there we can reasonably construct cycles since the rational resolution of the dilemmas and paradoxes lies in the ability to prefer an argument on the basis of information not included in a standard *AF*.

If, however, we consider polychromatic cycles, we get a different picture. If the cycle contains arguments of two different values, call them *red* and *blue*, at some point in the cycle a red argument must have a blue attacker, and a blue argument must have a red attacker. Now either red is preferred to blue, or vice versa: hence either the red argument attacked by the blue argument or the blue argument attacked by the red argument is not defeated. But if this is so, the preferred extension is not empty. Having established that one argument is in the preferred extension, we can now proceed round the cycle and determine the status of the remaining arguments. Similar reasoning can be applied to any polychromatic cycle. The actual status of an argument will, of course, depend on which argument was first decided to be in, which depends on the order of the values. We can, however, say that *for a given order on values* there will be a unique preferred extension. Thus we can say:
**Property 9**: *Given an order on values, a polychromatic cycle in a VAF has a unique, non-empty preferred extension.*
But cycles were the source of non-empty and multiple preferred extensions in *AFs*. See Dunne and Bench-Capon 2001 for a proof that an *AF* with more than one preferred extension must have an even cycle. Therefore we can say that a *VAF* containing no monochromatic cycles will as a whole have a unique, non-empty preferred extension, given an order on values. We can offer the following algorithm as a means of calculating this extension.

**Algorithm 10:** *EXTEND(AF,attacks)*.
1) $S := \{s \in AR: (\forall y)(not\ defeats(y,s))\}$
2) $R := \{r \in AR: \exists s \in S\ for\ which\ defeats(s,r)\}$
3) *If $S = \emptyset$ then return S and Halt*
4) $AR' := AR / (S \cup R)$
5) $attacks' := attacks / ((S \times R) \cup (R \times AF) \cup (AF \times R))$
6) *Return $S \cup EXTEND(AR',attacks')$*

The correctness of this algorithm is shown in Bench-Capon and Dunne (2002).

Note now the following: we can determine whether a dispute represented as a *VAF* is resoluble, and, given an order on values, credulous and sceptical acceptance are the same. The contents of the preferred extension, however, do depend on the value order. Now it is possible that some arguments may be in the preferred extension whatever the order on values and some in the preferred extension only for particular value orders. Let us call arguments in the preferred extension irrespective of value order *objectively acceptable*, those in the preferred extension for some value orders *subjectively acceptable*, and those in the preferred extension for no ordering on values *indefensible*. First, we may ask whether there are objectively acceptable arguments. Let us first consider the case of a dichromatic three-cycle. We may state that:

**Property 11**: *In a dichromatic three cycle the argument coloured differently from the other two will be objectively acceptable.*

**Proof:** First we define the notion of an *argument chain*.

**Definition 12**: An *argument chain* in a *VAF, C* is a set of n arguments $\{a_1 \ldots a_n\}$ such that:
   i.   $(\forall a)(\forall b)(a \in C\ \&\ b \in C) \rightarrow val(a) = val(b))$;
   ii.  $a_1$ has no attacker in C;
   iii. For all $a_i \in C$ if $i > 1$, then $a_i$ is attacked and the sole attacker of $a_i$ is $a_{i-1}$.

In an argument chain *C* it is obvious that, since all attacks will succeed because all arguments have the same value, if $a_1$ is accepted, then *every odd* argument of C is also accepted and *every even* argument of C is defeated. Similarly if $a_1$ is defeated, every odd argument of C is defeated and every even argument of C is accepted.

A three cycle will consist of a chain of two arguments with one colour (say red) and a chain of one argument of the other colour (say blue). Suppose blue greater than red. Now the blue argument is attacked by a red argument and so undefeated. Suppose on the other hand that red > blue. Thus the first argument in the red chain is undefeated and so the second argument is defeated. But this attacks the blue argument, which is therefore now defended by the first red argument and so undefeated. Thus the preferred extension contains the blue argument, irrespective of whether red > blue or blue > red. ÿ

As demonstrated in Bench-Capon and Dunne (2002) this result can be generalised for dichromatic odd cycles of any length to:

**Property 13**: *In a dichromatic odd cycle, the odd numbered arguments of any chain preceded by an even chain are objectively acceptable.*

Next consider even cycles. An even cycle must comprise either (1) an even number of odd chains, or (2) any number of even chains, or (3) a mixture of an even number of odd chains together with any number of even chains.

In case (1) the arguments in the preferred extension will depend on the ordering of the values. To be precise, it will comprise the odd numbered arguments from the chains with the preferred value and the even numbered arguments of the chains with the other value. This is observable from the fact that the first argument of the chain with the preferred value must be in, since the attack on it does not succeed. Hence all odd numbered arguments in that chain are also undefeated. In particular the last argument in the chain is not defeated, and so it defeats the first argument in the next chain, since its value is preferred. Hence the even numbered arguments in that chain will not be defeated.

In case (2) the preferred extension is independent of the ordering of the values, and will comprise the odd numbered arguments from each chain. The first argument of a chain with the preferred value will not be defeated. But this means that the last argument of that chain is defeated: hence the first argument of the succeeding chain is not defeated.

In case (3) the arguments in the preferred extension will depend on the value ordering, but the odd numbered arguments of any chain preceded by an even chain will be included irrespective of the value ordering. If the value of the odd chain is preferred, the first argument (and hence all odd numbered arguments) of this chain will be in the preferred extension. Suppose the value of the even chain is preferred. Now the first argument in this chain is not defeated since it must be attacked by the other, lesser, value. Hence the last argument will be defeated, since it is an even numbered member of a chain whose first argument is not defeated. Thus the first argument of the succeeding odd chain has a defence, and is not defeated.

So for even cycles we get:
**Property 14**: *In a dichromatic even cycle, the odd numbered arguments of any chain preceded by an even chain will be objectively acceptable.*

Note, however, that there may be no such argument, since the cycle may contain only odd chains. We may generalise these observations to:
**Property 15**: *In any dichromatic cycle, the odd numbered arguments of any chain preceded by an even chain will be objectively acceptable.*

But what of those chains preceded by odd chains? If the value of the preceding chain is preferred, the first argument will be defeated and the even numbered arguments will be in the preferred extension. If on the other hand the value of the preceding chain is not preferred, then the odd numbered arguments of this chain will be in. We are now in a position to characterise fully the preferred extension of a dichromatic cycle:
**Property 16**: *The preferred extension of a dichromatic cycle comprises:*
    (i)      the odd numbered arguments of all chains preceded by an even chain;
    (ii)     the odd numbered arguments of chains with the preferred value;
    (iii)    the even numbered arguments of all other chains.

Note that those included under (i) are objectively acceptable and those included under (ii) and (iii) are subjectively acceptable. The even numbered arguments of a chain preceded by an even chain are indefensible.

Now let us view such dichromatic cycles as part of a fuller dichromatic *VAF*. Consider the *VAF* in Figure 1.

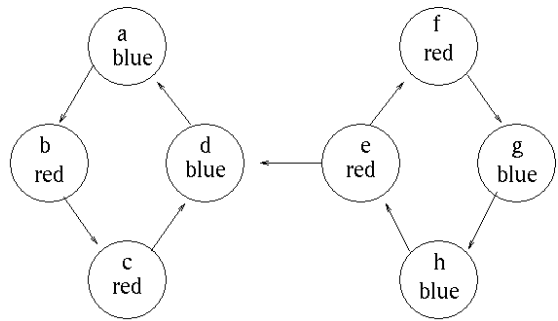

**Figure 1: Example *VAF***

Now the blue argument *d* is the successor not only of the chain *bc*, but also of the one argument chain *e*. (Argument *e* is part of two chains, *e* and *ef*). There will be two preferred extensions, according to whether red > blue, or blue > red. If red > blue, the preferred extension will be {*e,g,a,b*}, and if blue > red, {*e,g,d,b*}. Now *e* and *g* and *b* are objectively acceptable, but *d*, which would have been objectively acceptable if *e* had not attacked *d*, is only subjectively acceptable, and *a*, which is indefensible if *d* is not attacked, is also subjectively acceptable Arguments *c* and *h* remains indefensible. So, to be objectively acceptable, an argument must be an odd numbered member of a chain preceded *only* by even chains. Suppose that instead of *e* attacking *d*, *h* attacked *d*. In this case *d* would be part of the even chain *ghdb*, preceded by the even chain *ef*. Now {*egdb*) is the only preferred extension, whatever the value ordering, since all chains are preceded only by even chains.

The fact that arguments can be part of two chains complicates matters. To try to impose some order let us consider some cases. Suppose that an argument is an even numbered member of a chain which is unattacked or preceded only by even chains. Will it then be indefensible? This depends on what we consider a chain. Consider Figure 2.

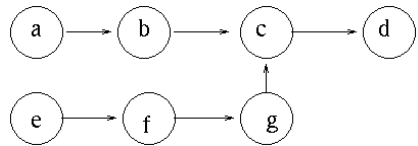

**Figure 2: Monochromatic *VAF* with argument attacked twice**

Suppose we see this as two chains, *abcd* and *efgcd*. Argument *d* is an even numbered member of the first. But in fact far from being indefensible, *d* is objectively acceptable, since *c* is always defeated because of its position in *efgcd*. In fact, we need to construe this situation as three chains: *abc*, *d* and *efgc*. Now *d* is no longer an even numbered member of a chain attacked only by even chains, since it is now odd numbered. Thus an argument with more than one attacker needs to be seen as the last element of any chain where the attackers are of its own colour, and its successors will form their own new chain. If a node terminates two or more chains, this is understood, when considering chains attacked by it, as an even chain if any of its chains are even. An argument can also be part of two chains if it attacks two arguments with the same value: in this case, however, it will either be odd numbered or even numbered for all such chains.

So, an argument is indefensible if it is an even numbered member of any chain attacked only by even chains (construing an unattacked chain as attacked by a chain of length zero). But there is another way that an argument can be indefensible. Consider Figure 3.

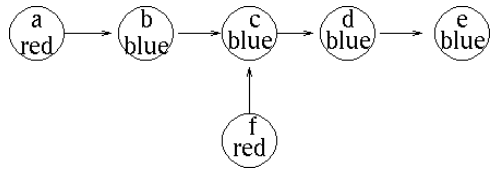

**Figure 3: Dichromatic *VAF* with an argument attacked twice**

Here we have the following chains: *a*, *bc*, *f*, *de*. If *f* was not present, we would have a single blue chain, *bcde*, and all the blue arguments would be subjectively acceptable: *b* and *d* if blue > red and *c* and *e* if red > blue. Adding *f*, however, changes this, since it splits the blue chain into *bc* and *de*. Now *de* is a chain preceded only by an even chain, hence *d* is objectively acceptable and *e* is indefensible. Argument *c* will also be indefensible, since it will be defeated by *b* if blue > red and by *f* if red > blue. This is best seen as construing *c* as forming a fifth chain, of which it is the only element. Now an argument will be indefensible if it forms part of two chains attacked by odd chains in which it is even numbered in one chain and odd numbered in the other, which can happen only if an even numbered member of a chain is *directly* attacked by an odd chain. Again where an argument terminates two or more chains, it is considered to be the end of an even chain if any of these chains are even, for the purposes of attacking succeeding chains.

So if an argument has two (or more) attackers it will form part of two chains. If the attacking argument is of the same colour it will be the last element of that chain; if the attacker is of a different colour it will form a single element chain. If an argument is the member of any even chain, any chains it attacks will be considered attacked by an even chain. The objectively acceptable arguments will be the odd numbered members of chains preceded only by even chains.

We can summarise this :
**Property 17**: *In a dichromatic VAF:*
    i.   *an argument is indefensible if it an even numbered member of any chain preceded only by even chains; or if it an even numbered member of a chain attacked by an odd chain, and is directly attacked by an odd chain;*

ii. *an argument is objectively acceptable if it is only an odd numbered argument of a chain preceded only by even chains;*
iii. *an argument is subjectively acceptable otherwise.*

An unattacked argument is considered to be preceded by a chain of length zero, hence an even chain.

Dichromatic argument frameworks are of considerable importance since we can construe many disputes, such as that discussed in the next Section, as involving only two values. Some disputes, however, particularly in law, seem to involve more than two values. We should therefore ask if we can say anything about polychromatic *VAFs* in general. First we can say that a polychromatic *VAF* containing no monochromatic cycles will, given an ordering on values, have a unique preferred extension. Moreover the EXTEND algorithm remains applicable. We can, however, now construct examples of chains preceded by even chains in which the odd values are not objectively acceptable. Consider a seven cycle with three values, arranged as two blues, three reds, and two greens. Here the first and third red arguments will be objectively acceptable. But the blue chain also follows an even chain. The first argument in the blue chain will be defeated if green > red > blue (although undefeated for any other ordering). We now have a possible position in between objective and subjective acceptance: an argument may be acceptable with respect to a partial ordering on values, which means that parties can agree to accept some subjectively acceptable arguments even though they disagree as to the ranking of *some* values. Polychromatic *VAFs* give rise to a number of questions which require further work to explore. In the rest of the paper I will speak many of dichromatic *VAFs*.

## 4. An Example of Disputes in the Moral Domain.

The scenario we will consider is taken from an example discussed in Coleman (1992), and further discussed in Christie (2000). Hal, a diabetic, loses his insulin in an accident through no fault of his own. Before collapsing into a coma he rushes to the house of Carla, another diabetic. She is not at home, but Hal enters her house and uses some of her insulin. Was Hal justified, and does Carla have a right to compensation? A diagrammatic representation of the *VAF* for this dispute is given as Figure 4.

The first argument is that Hal is justified, since a person has a privilege to use the property of others to save their life - the case of necessity. But should Hal compensate Carla? His justification can be attacked by an argument that if one infringes the property rights of another one should pay compensation. The first argument (a) is based on the value that life is important, the second (b) on the value that property owners should be able to enjoy their property. By valuing life over property we can accept both arguments: Hal was justified in taking the insulin, but is obliged to compensate Carla. This appears to be Coleman's view. But now we may add a third argument, (c) attacking (b). This says that even if Hal were too poor to compensate Carla, then he should still be able to take the insulin: no one should die through their poverty. This in turn could be attacked by argument (d) which is based on the fact that starvation is not a recognised defence against theft, even of food. This argument is itself attacked by (a), so we have a four cycle with alternating values. In this case, as we have seen, we can choose either of two consistent positions, depending on which value we prefer. Either we can value human life over property, in which case we reject (b) and (d), and do not oblige Hal to compensate Carla, or we reverse our preference and have Hal in default if he cannot pay the compensation.

But suppose we argue the case differently. Instead of using argument (b) we attack (a) with an argument (e) to the effect that Hal is endangering Carla's life by consuming her insulin. On her return she may be in need of the insulin which is no longer available. Here both (e) and (a) are based on the same value, so (e) defeats (a). The threat to Carla can be removed by Hal replacing the insulin, so we can attack (e) by an argument that Hal should replace the insulin, effectively argument (b) above. Note that (b), which says that Hal is justified if and only if he compensates, can be used both to demand compensation and so attack (a) and to remove the threat to Carla, so attacking (e). We now plead Hal's poverty and attack (b) using (c). Now, however, instead of attacking (c) with a property based argument as above, suppose we use a life based argument, (f) to the effect that no one should preserve their own life at the cost of endangering another's. This argument is attacked by (a) which held that one may do anything to preserve one's own life. This framework now contains a five cycle (a,e,b,c,f,a) with a chain of four life based arguments, and one property based argument (b). In such a situation *property 13* shows that we have objectively to accept the property based argument, and conclude that Hal is not justified in taking the insulin without compensation - whichever value we prefer. Even if we rate life more highly than property, we must conclude that Hal should not value his own life over Carla's, and so it does seem fair that the first claim on the insulin should go to its rightful owner rather than anyone who can get possession of it.

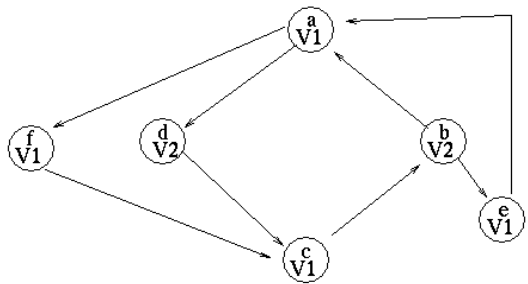

**Figure 4:** *VAF* **for Hal and Carla case: V1 is life, V2 is property**

Now consider the full *VAF* with all six arguments, shown in Figure 4. Suppose V1 (life) is preferred to V2 (property). Now (e) is in the preferred extension because it is attacked only by (b) which has the inferior value. So (a) is out, because attacked by (e), so (f) and (d) are in, because attacked only by (a). Now (c) is defeated by (f), and (b), attacked only by (c), is in. Thus, given V1 > V2, the rational position is $\{e,f,d,b\}$. Now suppose V2 > V1: now (b) is in, because it is only attacked by the lower ranked (c). Therefore (a) and (e) are defeated by (b). With (a) out, (f) and (d) are in and (f) defeats (c). So, for V2 > V1, the rational position is $\{b,f,d\}$. Thus (b), (f) and (d) are objectively acceptable, and (e) is subjectively acceptable, if we value life over property. The conclusion is that Hal can take the insulin *only* if he replaces it before Carla needs it, hence only if he is able to replace it. Note that (a) and (c) are indefensible: thus it looks as if the poverty defence is a non-starter, as is the proposition that one may do anything (even endanger the life of an innocent person) to preserve one's life.

But this is so only because of the availability of (e) in the framework. Suppose in fact Carla was a pharmacist, and not a diabetic. Now Carla's life is not endangered by Hal's action, since Carla has no need of insulin. Removing (e) from the framework gives two even cycles. Now the choice of preferred value determines which of $\{a,c\}$ or $\{b,d,f\}$ is the rational position. Note that (b) is no longer objectively acceptable, so that Hal can take the insulin without compensation, given a preference for V1 over V2 Although (f) may be in general desired, it need not be included in the position of someone who prefers life over property, since without (e) it is irrelevant. With respect to (d), such a value order would seem to license theft by a starving person - provided the theft was not from another starving person. We may object to this and wish to analyse further.

The further analysis might run as follows. Suppose we attack (a) by an argument (g) saying that the life of some other diabetic is endangered, since they may rely on Carla being in a position to supply them with insulin. We may attack this by a factual argument (h) to the effect that Carla is well stocked with insulin and will be able to meet any foreseeable demand. But we may again argue (k) that Hal cannot possibly know that this so: Carla may have allowed her stocks to run down. This gives the a *VAF* shown in Figure 5.

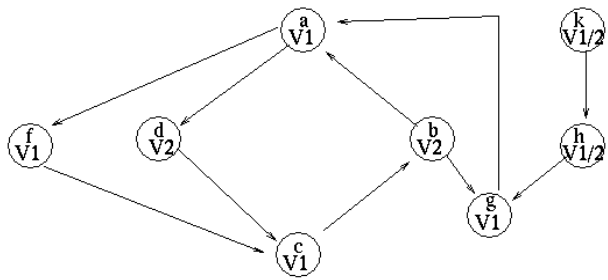

**Figure 5:** *VAF* **for Hal and the extended chemist scenario**

Note that a factual argument is taken as value free: it should therefore be capable of defeating any other argument irrespective of value. We model this within a dichromatic *VAF* by allowing it to take on the preferred value. Suppose here we allow it to take V1, since that will ensure that it always defeats (g). Now we have seven chains: *a, fc, khga, c* and *g* relating to V1 and *b* and *d* relating to V2. Now (a) and (c) are even members of an unattacked chain, and so indefensible, allowing all of (f), (b) and (d) to be objectively acceptable. If we prefer property to life, (g) will be rejected, otherwise it can be accepted. My contention is that this can be seen as representing the reasonable position in the dispute.

## 5. Strategic Heuristics in *VAF*s

Given a *VAF* we can determine the status of an argument. But suppose we do not like the status so determined. We need not concede at this point: we can extend the *VAF* by adding another argument. But, of course, we want to extend the *VAF* at a point where it will have the impact we desire. The considerations above, suggest some heuristics for extending the *VAF* to produce a particular status for a given argument.

The status of an argument depends on its position in chains of which it is part, and on the parity of the chains attacking those chains. A decision graph showing the dependency of status on these factors is shown in Figure 6.

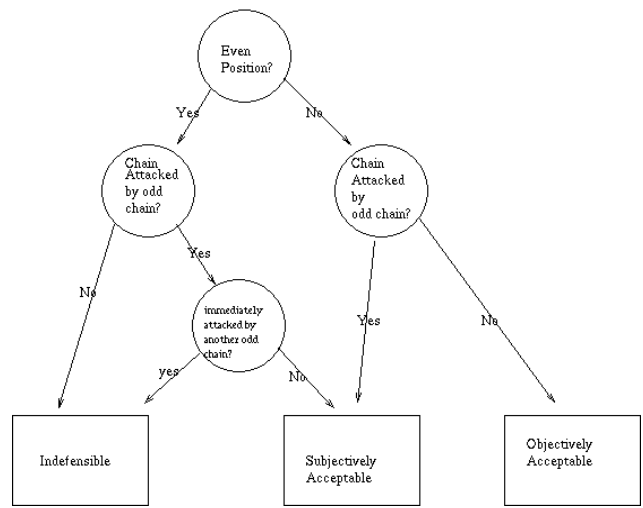

**Figure 6: Decision process for status of an argument in a dichromatic *VAF***

First note that the availability of this decision process means that we need consider only the position of an argument within its chains, and the parity of chains attacking its chains. No attack on a successor argument, or any more remote argument need be considered.

To change the position of an argument within a chain we can either extend the chain (with an argument of the same colour), if the chain is unattacked; or attack some preceding *odd numbered* argument in the chain (with an argument of either colour). To make an argument part of an additional chain we must attack that argument itself (again with an argument of either colour). We can change the parity of an attacking chain either by extending that chain (with argument of the colour of that chain), or by attacking some odd numbered argument in that chain (with an argument of either colour). Finally if a chain is attacked only by even chains, we may prevent this by attacking the head of that chain with an argument of different colour.

If an argument is objectively acceptable it is an odd numbered member of a chain attacked only by even chains. To make it indefensible we must attack it, or an odd numbered predecessor, with an argument of the *same* colour. To make it subjectively acceptable we must change the parity of one of its attacking chains, by attacking an odd numbered argument of the attacking chain with an argument of *either* colour, or adding an attacker by attacking the head of its chain with an argument of *different* colour.

If an argument is indefensible, it must be an even numbered member of a chain attacked only by even chains. To render it objectively acceptable we must attack an odd numbered predecessor with an argument of the *same* colour. It can be made subjectively acceptable by attacking an odd numbered argument of that chain with an argument of *either* colour, or adding an attacker by attacking the head of the chain with an argument of *different* colour, giving its chain an odd attacker.

If an argument is subjectively acceptable, it must be part of a chain attacked by at least one odd chain. If even numbered it can be made objectively acceptable by attacking an odd numbered predecessor with an argument of the *same* colour. It can be made indefensible by making all the attackers of its chain of even length, by attacking odd numbered members of them with arguments of *either* colour. Alternatively it can be made indefensible by attacking it, or an even numbered predecessor, with an argument of *either* colour. If it is odd numbered it can be made indefensible by attacking it, or an odd numbered predecessor with an argument of the *same* colour, and it can be made objectively acceptable by making all chains attacking its chain of even parity, by attacking odd numbered members of those chains with arguments of *either* colour.

With this information we can focus our attention on which arguments can be usefully attacked to produce the desired effect, and whether they need to be attacked by an argument of the same or different colour. These heuristics only consider the case where the *VAF* is extended by introducing a new argument attacking one other argument, which is probably the most usual way of continuing a dispute. Similar analysis could be used to provide heuristics if it is permitted to introduce a new attack between existing arguments, or to introduce several new arguments and attacks.

## 6. Concluding Remarks

In this paper I have introduced the notion of a Value-Based Argumentation Framework (*VAF*), an augmentation of standard Argumentation Frameworks, which represent an established way of exploring logics for defeasible reasoning. The key idea here is to allow for attacks to succeed or fail, depending on the relative worth of the values promoted by the competing arguments. *VAFs* have their basis in the important work of Perleman, who has persuasively argued that rationality need not be confined to demonstrable proofs, and that by recognising that arguments derive their force from promoting or protecting values shared by the audience of an argument, we may rationally choose amongst conflicting arguments, and that rational disagreement can be explained in terms of different priorities of the values associated with arguments. *VAFs* thus represent a way of adding this information to *AFs* so as to allow for this kind of explanation.

The main result of this paper is to show that in a *VAF* it may be possible to *force* rational acceptance of particular arguments within the *VAF*, *irrespective of how the values are ranked*. The importance of this is that it demonstrates how some arguments are objectively compelling, even in the face of different subjective preferences.

These ideas have been illustrated by exploring the properties of *VAFs*, in particular for *VAFs* restricted to two values, which can represent an important class of moral and legal disputes. The analysis that is possible with a *VAF* has been illustrated by a discussion of a well known moral dispute. Finally I have shown how the properties of *VAF*s can be used to produce heuristics for extending a dispute in order to achieve some particular position.